\pdfoutput=1

\documentclass[11pt]{article}

\usepackage[table]{xcolor} 
\usepackage[final]{acl}
\usepackage{tabularx}%
\usepackage{multirow}
\usepackage{times}
\usepackage{latexsym}
\usepackage{booktabs}
\usepackage{arydshln}

\usepackage{placeins}

\usepackage{hyperref}
\usepackage[T1]{fontenc}

\usepackage[utf8]{inputenc}

\usepackage{microtype}

\usepackage{inconsolata}

\usepackage{graphicx}
\usepackage{multirow}
\usepackage{subfig}
\usepackage{color}
\usepackage{colortbl}
\usepackage{multirow}
\usepackage{CJKutf8}
\usepackage{dsfont}
\usepackage{textcomp}

\makeatletter
\def\adl@drawiv#1#2#3{%
        \hskip.5\tabcolsep
        \xleaders#3{#2.5\@tempdimb #1{1}#2.5\@tempdimb}%
                #2\z@ plus1fil minus1fil\relax
        \hskip.5\tabcolsep}
\newcommand{\cdashlinelr}[1]{%
  \noalign{\vskip\aboverulesep
           \global\let\@dashdrawstore\adl@draw
           \global\let\adl@draw\adl@drawiv}
  \cdashline{#1}
  \noalign{\global\let\adl@draw\@dashdrawstore
           \vskip\belowrulesep}}
\makeatother

\title{ECoh: Turn-level Coherence Evaluation for Multilingual Dialogues}

\author{John Mendonça\textsuperscript{1,2}, Isabel Trancoso\textsuperscript{1,2} \and  Alon Lavie\textsuperscript{3,4}\\
  \textsuperscript{1} INESC-ID, Lisbon \\
  \textsuperscript{2} Instituto Superior Técnico, University of Lisbon \\
  \textsuperscript{3} Carnegie Mellon University, Pittsburgh \\
  \textsuperscript{4} Phrase, Pittsburgh \\
  \texttt{\{john.mendonca, isabel.trancoso\}@inesc-id.pt}, \texttt{alavie@cs.cmu.edu} \\}

\begin{document}
\maketitle

\begin{abstract}

Despite being heralded as the new standard for dialogue evaluation, the closed-source nature of OpenAI's GPT-4 model poses challenges for the research community. Motivated by the need for lightweight, open source, and multilingual automated dialogue evaluators, this paper introduces \textsc{GenResCoh} (\textbf{Gen}erated \textbf{Res}ponses targeting \textbf{Coh}erence).  \textsc{GenResCoh} is a novel LLM-generated dataset comprising over 130k negative and positive responses and accompanying explanations seeded from XDailyDialog and XPersona covering English, French, German, Italian, and Chinese. Leveraging \textsc{GenResCoh}, we propose \textsc{ECoh}\footnote{Pronounced \textit{"Echo"}.} (\textbf{E}valuation of \textbf{Coh}erence), a family of evaluators trained to assess response coherence across multiple languages. Experimental results demonstrate that \textsc{ECoh} achieves multilingual coherence detection capabilities superior to the teacher model (GPT-3.5-Turbo) on \textsc{GenResCoh}, despite being based on a much smaller architecture. Furthermore, the explanations provided by \textsc{ECoh} closely align in terms of quality with those generated by the teacher model.

\end{abstract}

\section{Introduction}

\begin{table}[t]
\small
\begin{tabularx}{0.48\textwidth}{ X }
\toprule
\textbf{Context:}\\
\textbf{A:} Hello. I bought a China dress in your shop this morning.\\
\textbf{B:} Yes?\\
\textbf{A:} I bought it one size up by mistake.\\
\textbf{B:} Oh, did you?\\
\textbf{A:} I wonder if you can change it to one size down.\\\midrule
\textbf{Reference Response.} \textbf{B:} Yes, of course. Will you come with the receipt ?\\
\cellcolor[HTML]{ECF9E9}\textit{"The response acknowledges the request and offers a solution to accommodate the customer's needs. The answer is Yes."}\\ 
\midrule
\textbf{Random Negative Sample.} \textbf{B:} I'd like a book about law.\\ 
\cellcolor[HTML]{FDEDEC}\textit{"The response is completely unrelated to the situation discussed. The answer is No."}\\
\midrule
\textbf{\textsc{GenResCoh} Positive Sample.} \textbf{B:} Absolutely, if you bring the dress back to the shop with the receipt, we can exchange it for a smaller size for you.\\
\cellcolor[HTML]{ECF9E9}\textit{"The response acknowledges the request and offers a solution to the problem. The answer is Yes."}\\ \midrule
\textbf{\textsc{GenResCoh} Negative Sample.} \textbf{B:} Oh, that’s great to hear! I hope it fits perfectly.\\
\cellcolor[HTML]{FDEDEC}\textit{"The response does not acknowledge the request for a size change and instead expresses an unrelated sentiment. The answer is No."}\\
\bottomrule
\end{tabularx}
\caption{Example of automatically generated negative samples obtained with random response selection, and obtained from \textsc{GenResCoh} (\S\ref{sec:negeval_dataset}), our proposed dataset. The explanations are generated using one of our proposed models, \textsc{ECoh-4B-ml} (\S\ref{sec:method}).}
\label{tab:sampleshowcase}
\vspace{-0.65cm}
\end{table}

With LLMs showcasing impressive reasoning and dialogue understanding capabilities vastly superior to any prior NLP technologies, human evaluation has more recently been complemented with automatic evaluations using GPT-4 \citep{openai2024gpt4}. However, GPT-4 as an automated evaluator has its downsides. Perhaps the main downside is it being a closed source model hidden behind a paid API, making accessibility difficult for those outside the coverage area and lacking extensive financial resources, while also lacking transparency in its development.
In contrast, and to the best of our knowledge, the study of open source alternatives to GPT-4 based dialogue evaluation is mostly limited to the benchmarking of open source and open access LLMs or finetuning with dialogue data \cite{huynh2023understanding, zhang-etal-2023-xdial, zhang2024comprehensive}. These works suggest that LLMs struggle to outperform older encoder-based metrics trained using negative sampling approaches for relevance (e.g. random response selection). However, it is important to point out that these benchmarks have several limitations.

First and foremost, the high performance of these encoder-based models can be explained by the fact that the benchmarks themselves are based on old generative models that exhibit relevance issues that are easy to detect. For instance, in Table \ref{tab:sampleshowcase}, metrics trained using random negative sampling strategies for relevance will output a positive score to all responses except the random negative one. As such, these metrics struggle to evaluate contemporary chatbots, since these typically output fluent and semantically relevant responses.

Furthermore, only a select few benchmarks are multilingual. Whilst there is work that attempts to evaluate the multilingual capabilities of dialogue evaluation metrics \cite{mendonca-etal-2023-towards, zhang-etal-2023-xdial}, they use translated benchmarks. This assumes that critical errors typically produced by these older models (e.g. irrelevance), are not influenced by language. However, more complex quality aspects such as coherence may have nuances that make them unique to certain cultures. Depending on the context, some culture specific details may or may not be implicitly inferred \citep{Hall1959language}.

These key observations motivate our work. In order to move towards the development of metrics that evaluate dialogue coherence and are multilingual, we propose \textsc{GenResCoh} (\textbf{Gen}erated \textbf{Res}ponses targeting \textbf{Co}herence), a collection of positive and negative responses focused on coherence. Our dataset, generated using strong LLMs, contains over 130k responses in different languages (English, French, German, Italian, and Chinese), together with their corresponding explanations (in English). By prompting an LLM, we are able to (1) obtain positive samples that are in distribution (LLMs frequently output more verbose responses than their human counterparts); (2) obtain negative samples that remain semantically relevant but contain coherence and logical consistency issues, which may be more informative during training, and that are more representative of current limitations of LLMs.

With this dataset, we train a family of evaluators we call \textsc{ECoh} (\textbf{E}valuation of \textbf{Coh}erence)\footnote{\url{github.com/johndmendonca/Ecoh}}. Our results demonstrate that distilling Coherence knowledge from a strong LLM allows us to obtain multilingual coherence detection performance of \textbf{.945} F1 score using a 0.5B model, which is superior to both the teacher models' (GPT-3.5-Turbo) \textbf{.910} and a much larger model of the same family (\textsc{Qwen1.5-7B-Chat} - \textbf{0.825}). Furthermore, the explanations provided by \textsc{ECoh} are of higher quality than \textsc{Qwen1.5-7B-Chat}, scoring an average of over \textbf{4} out of 5 on most instances, as reported by GPT-4 evaluations.

\begin{table*}[t]
\small
\centering
\begin{tabular}{lcccc}
\toprule
\textbf{Dataset} & \textbf{Size (\# contexts)}         & \textbf{Response Avg. length}     & \textbf{Explanation Avg. length} & \textbf{Response MTLD}             \\ \midrule
DailyDialog++ \citeyearpar{sai-etal-2020-improving}& & & &\\\cmidrule{1-1}
\hspace{1.25cm} Random      & \multirow{2}{*}{9,259/1,028/1,142}            & 9.40                         & -                         & 169.94                              \\               
\hspace{1.25cm} Adversarial &                              & 10.70                        & -                         & 186.42                              \\ \midrule\midrule
\textsc{GenResCoh-dev}& & & &\\\cmidrule{1-1}
\hspace{0.3cm} DailyDialog-\textsc{latin}            & \multirow{2}{*}{51,873/5,080} & 14.74                        & 15.03                     & 105.03                               \\
\hspace{0.3cm} DailyDialog-\textsc{zh}            &                               & 23.06                        & 14.54                     & 54.38                               \\ \midrule
\textsc{GenResCoh-test} & & & &\\\cmidrule{1-1}
\hspace{0.3cm} DailyDialog-\textsc{latin}           & \multirow{2}{*}{4,770}  & 14.82                            & 26.27                     & 155.28                              \\
\hspace{0.3cm} DailyDialog-\textsc{zh}           &                               & 24.79                        & 26.04                     & 69.03                               \\\cmidrule{2-5}
\hspace{0.3cm} PersonaChat-\textsc{latin}           & \multirow{2}{*}{1,000}        & 15.37                        & 27.89                     & 204.61                             \\
\hspace{0.3cm} PersonaChat-\textsc{zh}           &                               & 28.81                        & 27.78 & 76.66                              \\\bottomrule
\end{tabular}
\caption{Comparison of statistics for different negative sample datasets. DD denotes XDailyDialog, PC XPersona. Dataset size denotes the number of unique contexts from which responses were obtained for training/validation/test subsets. MTLD denotes the Measure of Textual Lexical Diversity \citep{mccarthy2005assessment} of the responses. We report statistics for Latin script languages (denoted \textsc{latin} and covering \textsc{en},\textsc{de},\textsc{fr},\textsc{it}), separated from Chinese-\textsc{zh}. For Average length, \textsc{latin} is calculated using words, whereas \textsc{zh} uses characters.} 
\label{tab:data_stats}
\end{table*}

\section{\textsc{GenResCoh} responses dataset}
\label{sec:negeval_dataset}


This section introduces \textsc{GenResCoh}, a multilingual, large-scale response collection that targets coherence, seeded from well established dialogue datasets (\S \ref{sec:preliminaries}), and generated using LLMs (\S \ref{sec:collection_generation}). Table \ref{tab:sampleshowcase} presents an example from this dataset. For additional examples of this dataset in other languages, see Appendix \ref{sec:app_ex}. 

\subsection{Dataset Sources}
\label{sec:preliminaries}

Our work leverages two distinct dataset sources: XDailyDialog \citep{liu-etal-2023-xdailydialog} and XPersona \citep{lin2021xpersona}. For training, development and testing, we use XDailyDialog, a multilingual extension of DailyDialog with human translations covering German-\textsc{de}, Italian-\textsc{it} and Chinese-\textsc{zh}. XDailyDialog includes 13K parallel dialogues, amounting to 52K dialogues and 410K utterances. During our pre-processing step we noted a substancial overlap of dialogues between the provided test and training/validation sets of XDailyDialog. As a result, we excluded these dialogues (amounting to 20\%) from the test set.

In order to gauge the extensibility to other dialogue datasets and languages, we additionally include XPersona data in our \textsc{GenResCoh} test set. XPersona is a multilingual extension of the PersonaChat dataset \cite{zhang-etal-2018-personalizing} with human revised machine translations for six languages. Besides English, we include Italian-\textsc{it}, Chinese-\textsc{zh}, and an additional unseen language, French-\textsc{fr}, in our experiments. For each language, we extract 1K contexts from the test set for response generation.

For contrastive comparison, we also use DailyDialog++ \citep{sai-etal-2020-improving}, a similar curation effort which uses the original DailyDialog dataset, and where annotators were asked to create five additional relevant responses and five adversarial irrelevant responses for each context.

\subsection{Generation}
\label{sec:collection_generation}

\paragraph{Development set} We leverage GPT-3.5-turbo\footnote{\texttt{gpt-3.5-turbo-0125} and \texttt{gpt-4-1106-preview} accessed via OpenAI's API in early April.} \citep{ouyang2022training} as the strong LLM to generate, given prior dialogue context, a positive and a negative response, paired with a brief explanation of the issue (or lack thereof). 
Each response pair is generated given a context of at least 2 turns up until the length of the dialogue except the last turn (this ensures the response is generated from a still ongoing conversation). We set the temperature to 0.7, the top-$p$ to 1, and the maximum number of tokens to 300, thereby enforcing smaller explanations which in turn should reduce inference costs. Despite sharing the same contexts, the responses and corresponding explanations are not necessarily translations of the English subset. This allows the model to freely generate responses that are more likely to occur (for the positive samples) or more representative of coherence issues in that particular language, instead of being a translation from English. The prompt used for this generation is included in Appendix \ref{sec:app_generate}.

\paragraph{Test set} For testing, we employ GPT-4 \citep{openai2024gpt4} to ensure higher quality outputs and reduce in-distribution biases from the training set. GPT-4 has been shown to match human annotations on quality, from general NLP tasks to highly specialised fields \citep{west-etal-2022-symbolic, raunak-etal-2023-leveraging, 10.1145/3587102.3588792}.

\paragraph{Human validation} In order to verify the outputs of GPT-4, we additionally conduct a human validation step involving one expert linguist from each language. 
We randomly sample 100 examples from the XDailyDialog test set, and report an appropriateness rate that exceeds 97\%, thus validating the response and explanation generation process using GPT-4. Details regarding human validation are provided in Appendix \ref{sec:app_validate}.

\subsection{Statistical Analysis}

We present relevant statistics for our dataset, together with DailyDialog++ in Table \ref{tab:data_stats}. Since the test set for our dataset is generated by GPT-4, we opt to
present the statistics separately. 

Firstly, despite \textsc{GenResCoh} boasting a much larger context set, amounting to 51k/5k for training/validation, each context only has a single positive and negative response, whereas DailyDialog++ contains 5 positive responses and an additional 5 adversarial negative responses.

When comparing the average length of responses, we note that \textsc{GenResCoh} responses are longer than the human curated responses of DailyDialog++. This verbosity is a known behaviour by LLMs, since they are conditioned to output longer responses due to the Reinforcement learning from human feedback (RLHF) step, at least when compared to humans \citep{kamalloo-etal-2023-evaluating}. Additionally, we note that the response lengths remain similar across the development and test sets, whereas the explanations are much longer in the test set. 

For a more fine grained analysis of the responses, we measure their lexical diversity using the Measure of Textual Lexical Diversity (MTLD) \citep{mccarthy2005assessment}.\footnote{Calculated using \texttt{lexical-diversity} Python package.} Since DailyDialog++ contains 5 responses per context, we calculate the average diversity when considering the responses individually. We observe that the diversity of human responses for DailyDialog++ is larger than the ones generated by GPT-3.5-Turbo for the development set, but similar to the ones generated by GPT-4 for the test set. This disparity is to be expected, given the performance differences between the two models in creative writing tasks.\footnote{It is important to point out, that a higher temperature value would likely result in higher diversity, with a possible trade off in performance.}

It is important to note that the adversarial responses from DailyDialog++ exhibit greater diversity compared to those from \textsc{GenResCoh}. This is because the tasks are slightly different: in DailyDialog++, annotators were asked to generate new irrelevant responses by incorporating certain words from the context directly or indirectly into their responses. This stands in contrast to our approach, which prioritises coherence while preserving relevance. As such, the introduction of diverse words into the response is constrained by the fact relevance must be uphold.

\section{\textsc{ECoh}}
\label{sec:method}

This section presents \textsc{ECoh}, our proposed family of response coherence evaluators. We initially present the method of formulating the task of coherence evaluation as explainable QA (Question Answering) (\S\ref{sec:problem_formulation}). Then, we describe in detail how our evaluator is trained (\S\ref{sec:exp_setup_main}) and evaluate its performance on different settings (\S\ref{sec:main_res}).

\subsection{Problem Formulation}
\label{sec:problem_formulation}

Turn-level dialogue coherence evaluation consists of the assessment of a response hypothesis $h$ given a dialogue history (frequently denoted as context) $c$ of varying amount of turns, and optionally one or more references $r$ and/or external knowledge $k$. The goal is to learn a scoring function that assigns a score $f(c,k,r,h) \rightarrow s$ for each individual quality aspect. This scoring function is compared against human judgements, which annotate the same context-response pairs. These responses are evaluated by humans using, for instance, a binary $(0,1)$ judgement or a $[1,5]$ Likert scale, where the lowest value means lowest quality and highest value maximum quality. 

In our work, we consider Coherence as being a binary quality aspect. Despite being frequently annotated in the literature on a Likert Scale, what can be considered a response that is neither coherent or incoherent is mostly left to the interpretation of the annotator. Given that we are leveraging an LLM for generation, we find it unfeasible to generate a balanced dataset that contains intermediate levels of coherence. Instead, we generate a positive and a negative response in terms of coherence and label it accordingly. This constrastive sampling strategy for coherent responses is also followed in most metric development work for Relevance or Sensibleness, where models are typically trained using self-supervised learning strategies that sample negative responses by random selection \citep{mehri-eskenazi-2020-usr, yeh-etal-2021-comprehensive, mendonca-etal-2023-simple}. Lacking any external knowledge with respect to each dialogue, we then further simplify the reference-free evaluation of coherence as a Question Answering (QA) task ($f(c,h) \rightarrow s \in (0,1)$), with model responses being either coherent ("Yes") or incoherent ("No").

\subsection{Experimental Setup}
\label{sec:exp_setup_main}

\paragraph{Model Specification} We employ the \textsc{Qwen1.5-Chat} family of LLMs \citep{qwen} for our models. \textsc{Qwen1.5} contains LLMs of various sizes, ranging from 0.5B up to 72B and support all the languages of XDailyDialog. We limit our finetuning experiments up to 4B due to the tradeoff between performances and compute. We feed the dialogue context to the model and ask it to provide a "Yes"/"No" answer to the question \textit{"Given the context, is the response Coherent?"}. The model is trained to also output a succinct explanation to the answer. We opted with asking for the explanation first, before answering the question, in order to leverage the autoregressive nature of the model. In theory, this should guarantee that final answer be informed by the explanation.\footnote{\citet{chiang-lee-2023-closer} has shown that dialogue evaluation performance is not \textit{always} better when requesting the explanation first. We leave this analysis for future work.} Additional training details are available in \ref{sec:exp_details}.

\paragraph{Baselines} We contrastively compare our proposed approach against several models. We begin by including models trained using random negative responses from DailyDialog: a \textsc{RoBERTa-large} model \citep{liu2019roberta} (which we train ourselves -- see Appendix \ref{sec:exp_details} for details); and \textsc{UniEval} \citep{zhong-etal-2022-towards} (which uses T5 as base model). Since these models output a probability score, we assume the model outputs the positive class when the $p>0.5$.  Additionally, we conduct zero and one shot (with English and language specific examples) inference using \textsc{Qwen1.5-Chat} to determine if finetuning on \textsc{GenResCoh} adds improvements to the performance of the base model. We also compare against GPT-3.5-Turbo \citep{ouyang2022training}, the teacher model which was used to generate the development set of \textsc{GenResCoh}, and which is weaker than our expert (GPT-4).

\subsection{Main Results}
\label{sec:main_res}

Since the coherence labels are binary, we report detection results using F1-score and Point Biserial Correlation. Additionally, we compute the BLEU-4 score of the generated short explanation using the GPT-4 explanation as a reference. Since BLEU compares overlap in tokens instead of comparing meaning, we also employ GPT-4 as a drop-in replacement for human annotators, and ask it to assess the explanations of 200 random responses from the models that output an explanation.

\begin{table}[ht]
\small
\centering
\begin{tabular}{lcccc}
\toprule
\textbf{Model}              & $\rho_{pb}$ & \textbf{F1}    & \textbf{BLEU}  & \textbf{GPT-4} \\\midrule
$\mathds{1}$ (always positive)                   & NaN       & .333     & -     & -     \\\midrule
\textsc{NSP-RoBERTa}           & .1651    & .430     & -     & -     \\
\textsc{UniEval}                 & \underline{.3272}         & \underline{.500}     & -     & -     \\\midrule
\multicolumn{5}{l}{\textsc{Qwen1.5-Chat}}\\\cmidrule{1-1}
\hspace{1.5cm}0.5B       & .2226   & .600 & 3.80  & 1.84\tiny{\textpm1.12}     \\
\hspace{1.5cm}1.8B       & .5212   & .740 & 2.58     & 2.39\tiny{\textpm1.29}     \\
\hspace{1.5cm}4B         & .5850   & .783 & \underline{8.16}     & 3.18\tiny{\textpm1.60}     \\
\hspace{1.5cm}7B         & \underline{.7918}   & \underline{.890} & 4.63     & \underline{3.95\tiny{\textpm1.48}}     \\\midrule
GPT-3.5-Turbo      & .8256        & .910     & 5.25     & \textbf{4.55\tiny{\textpm1.08}}     \\\midrule
\multicolumn{5}{l}{\textsc{ECoh-en}}\\\cmidrule{1-1}
\hspace{1.5cm}0.5B & .7756   & .878 & 16.02 & 3.80\tiny{\textpm1.43}     \\
\hspace{1.5cm}1.8B & .8242   & .908 & 17.30 & 4.13\tiny{\textpm1.29}\\
\hspace{1.5cm}4B & .9185   & .960 & 17.92 & \underline{4.45\tiny{\textpm0.96}}     \\
\multicolumn{5}{l}{\textsc{ECoh-ml}}\\\cmidrule{1-1}
\hspace{1.5cm}0.5B                 & .8882   & .945 & 17.00 & 3.99\tiny{\textpm1.36}     \\
\hspace{1.5cm}1.8B                 & .9019   & .953 & 17.28 & 4.24\tiny{\textpm1.28}\\
\hspace{1.5cm}4B & \textbf{\underline{.9491}}    &\textbf{\underline{.975}} & \textbf{\underline{18.05}}     & 4.29\tiny{\textpm1.12}    \\\bottomrule

\end{tabular}
\caption{Reported results on \textsc{GenResCoh-DD-test}, averaged across all languages. $\rho_{pb}$ denotes Point Biserial Correlation. \textsc{ECoh-en} and \textsc{ECoh-ml} denote the finetuned models using English data and all multilingual data, respectively. All correlation results are $p<0.05$. \textbf{Bold} denotes best overall model, \underline{underline} best model of the group.}
\label{tab:main_res}
\end{table}

We collate our main results in Table \ref{tab:main_res}. Due to space limitations, we only report 1-shot performance with a language specific example for \textsc{Qwen1.5-Chat} and the results correspond to the average of the languages. Additional results, including Zero shot and individual language performance, are available in Appendix \ref{sec:app_res}.

\paragraph{GPT-3.5 performance with 4B parameters} Our main observation is that, although being one of our smallest models, \textsc{ECoh-0.5B-ml} outperforms the predictive performance of the teacher model (reported in F1), and the explanations of \textsc{Qwen1.5-Chat}-7B. Furthermore, \textsc{ECoh-4B-en} has similar explanation quality to that of GPT-3.5-Turbo. As expected, training models using random response selection (\textsc{NSP-RoBERTa-l} and \textsc{UniEval}) is not sufficient for accurately detecting more advanced coherence issues. In fact, these models' performance sit between \textsc{Qwen1.5-0.5B-Chat} (.600 F1) and the naive single output model (.333 F1). 

\paragraph{Model size and Multilingual finetuning} Since our smallest model already achieves strong results (.945 F1 score), increasing the model size results in only a small relative improvement of 3\% in performance. However, we do observe larger performance improvements with multilingual finetuning. For instance, for \textsc{ECoh-0.5B}, we observe an improvement of over 7\% (.878 to .945). This indicates, as expected, that including multilingual data during finetuning improves results for the various covered languages. 

\paragraph{Explanations} We also note that our finetuned models have much higher BLEU and GPT-4 scores than the base models. The obtained BLEU scores are to be expected, given that \textsc{ECoh} is finetuned with explanation data stemming from the same prompt, which is a biased observation from the response generator. This is supported by the teacher model's performance, achieving the highest GPT-4 assessment, despite having low BLEU. In any case, by validating the responses of the \textsc{ECoh} models with GPT-4, we see that the explanations are on average of higher quality than the ones generated by even the largest base model (\textsc{Qwen1.5-Chat}) that we studied.

\subsection{Generalization to unseen dialogue datasets and languages}

In order to evaluate our models' capabilities on unseen dialogue datasets, we evaluate our models on XPersona, which was not seen during finetuning. We only select the best baselines (as reported in Table \ref{tab:main_res}) for this analysis. Additionally, our XPersona subset contains French, which is not present in XDailyDialog, so in addition to the average performance across all languages, we present the results for French separately. For fair comparison, we utilise the English example when evaluating the performance of \textsc{Qwen1.5-Chat} in French.

\begin{table}[ht]
\small
\centering
\begin{tabular}{lcccc}
\toprule
\textbf{Model}              & $\rho_{pb}$ & \textbf{F1}    & \textbf{BLEU}  & \textbf{GPT-4} \\\midrule
\multicolumn{5}{l}{\textsc{Qwen1.5-Chat-7B}}\\\cmidrule{1-1}
\hspace{1.5cm}\textsc{fr}   & .4608       & .660     & 2.97     & 3.20\tiny{\textpm1.58}        \\
\hspace{1.5cm}\textsc{ml}   & .6125       & .778     & 3.43     & 3.75\tiny{\textpm1.51}        \\\midrule
\multicolumn{5}{l}{\textsc{GPT-3.5-Turbo}}\\\cmidrule{1-1}
\hspace{1.5cm}\textsc{fr} & .7205   & .860  & 5.04  & 4.32\tiny{\textpm1.25}      \\
\hspace{1.5cm}\textsc{ml} & .7631   & .880  & 4.94  & \textbf{4.45\tiny{\textpm1.05}}      \\\midrule
\multicolumn{5}{l}{\textsc{ECoh-ml}}\\\cmidrule{1-1}
0.5B& & & & \\
\hspace{1.5cm}\textsc{fr} & .8089   & .910 & 13.71 & 3.68\tiny{\textpm1.46}     \\
\hspace{1.5cm}\textsc{ml} & .8882   & .945 & \textbf{\underline{17.00}} & 3.82\tiny{\textpm1.38}     \\
1.8B& & & & \\
\hspace{1.5cm}\textsc{fr} & .7863   & .890 & 14.10 & 4.40\tiny{\textpm1.15}     \\
\hspace{1.5cm}\textsc{ml} & .8472   & .920 & 15.70 & 4.26\tiny{\textpm1.17}     \\
4B& & & & \\
\hspace{1.5cm}\textsc{fr} & .9270   & .960 & 14.58 & 4.36\tiny{\textpm0.95}     \\
\hspace{1.5cm}\textsc{ml} & \textbf{\underline{.9448}}   & \textbf{\underline{.970}} & 16.33     & 4.38\tiny{\textpm0.96}    \\\bottomrule
\end{tabular}
\caption{Reported results for \textsc{GenResCoh-PC-test} (French-\textsc{fr} subset and full-\textsc{ml} set). $\rho_{pb}$ denotes Point Biserial Correlation. All correlation results are $p<0.05$. \textbf{Bold} denotes best overall model, \underline{underline} best model of the group.}
\label{tab:persona_res}
\end{table}

Looking at the results in Table \ref{tab:persona_res}, we find that the conclusions from DailyDialog also carry over to XPersona. For the unseen language (French-\textsc{fr}), we note a large drop in performance for \textsc{Qwen1.5-Chat-7B}, when compared to the other languages, which could be explained by the 1-shot example being in English. For our proposed models, we see a larger gap in performance between French and the other languages for the smaller models, whereas for \textsc{ECoh-4B}, the performance for French is well within the range of that observed for other languages. This is also what we observe for \textsc{GPT-3.5-Turbo}. This finding suggests that, given an LLM that natively supports languages for which we have no finetuning data, coherence knowledge can be drawn from languages that were included for finetuning.\footnote{It is important to acknowledge that this finding is only likely to extend to languages that follow western normative rules for coherence. An additional interesting experiment would be to test \textsc{ECoh} on a language that does not conform to these rules -- however, these are typically low-resource.}

\subsection{Generalization to external annotations}

Since the models were trained and evaluated on synthetic data, it is important to check if \textsc{ECoh} performs adequately on external evaluations conducted by human annotators. As such, we also assess \textsc{ECoh} on the FED-turn annotations \citep{mehri-eskenazi-2020-unsupervised} for "Relevance" and "Overall", which is a typically used benchmark for dialogue evaluation. Similar to other works, we calculate the average human annotation ($[0,2]$ for Relevance and $[0,4]$ for Overall) and report results using Spearman correlation between the human annotation and the score provided by each evaluator. For the LLMs, we keep the binary formulation for coherence (score is either 0 or 1). For the coherence explanation evaluation, lacking a reference, we again use GPT-4 as an explanation evaluator but without a reference response, and evaluate all responses. In order to gauge evaluation performance, we also calculate correlations with GPT-4 as a response evaluator. We refrain from providing GPT-4 explanation scores due to potential self-evaluation bias.

\begin{table}[ht]
\small
\centering
\begin{tabular}{lcccc}
\toprule
\textbf{Model}         & Relevance $r$ & Overall $r$ & GPT-4\\\midrule
\textsc{NSP-RoBERTa} & .2530 & \underline{.2543} & -\\
\textsc{UniEval}        & \underline{.2532}  & .2521 & -\\\midrule
\multicolumn{5}{l}{\textsc{Qwen1.5-Chat}}\\\cmidrule{1-1}
\hspace{1.5cm}0.5B	& \textit{.0451}	& \textit{.0340} & 2.35\tiny{\textpm1.42} \\
\hspace{1.5cm}1.8B	& .2693	&.2228  & 2.91\tiny{\textpm1.52} \\
\hspace{1.5cm}4B	& .1613	& .1189 & 3.30\tiny{\textpm1.67} \\
\hspace{1.5cm}7B    & \underline{.3279}        & \underline{.2998}  & \textbf{\underline{3.74\tiny{\textpm1.54}}} \\\midrule
GPT-3.5-Turbo      & .4025    & .3636 & 3.54\tiny{\textpm1.66} \\
GPT-4              & \textbf{\underline{.5108}}        & \textbf{\underline{.5320}} & -\\\midrule
\multicolumn{5}{l}{\textsc{ECoh}}\\\cmidrule{1-1}
0.5B& & &  \\
\hspace{1.5cm}EN  & .2247   & .1548 & 3.17\tiny{\textpm1.77}    \\
\hspace{1.5cm}ML  & .1670   & .1294 & 3.17\tiny{\textpm1.77}    \\
1.8B& & & \\
\hspace{1.5cm}EN & \underline{.2941}   & .2408  & 3.38\tiny{\textpm1.77}   \\
\hspace{1.5cm}ML & .2581   & .1801  &  \underline{3.50\tiny{\textpm1.68}}  \\
4B& & &  \\
\hspace{1.5cm}EN & .2445   & .2326 & 3.17\tiny{\textpm1.82}    \\
\hspace{1.5cm}ML & .2685   & \underline{.2515} & 3.37\tiny{\textpm1.81}      \\\bottomrule
\end{tabular}
\caption{Reported results for FED-Turn. Performance is calculated using Pearson Correlation ($r$). All results are $p<0.05$ unless \textit{italicised}. \textbf{Bold} denotes best overall model, \underline{underline} best model of the group.}
\label{tab:fed_res}
\end{table}

\begin{figure*}
\centering
\includegraphics[width=0.325\textwidth]{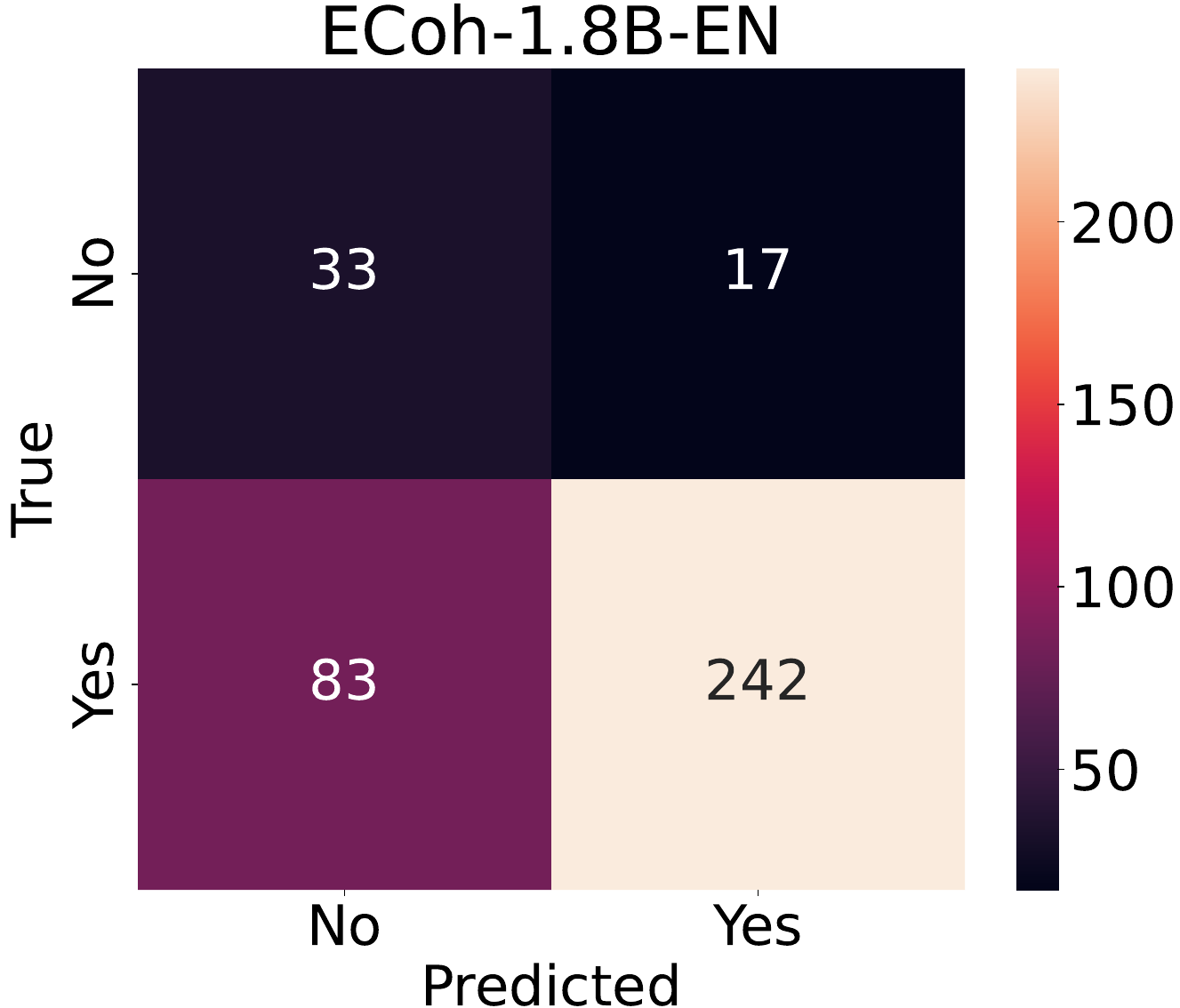}
\includegraphics[width=0.325\textwidth]{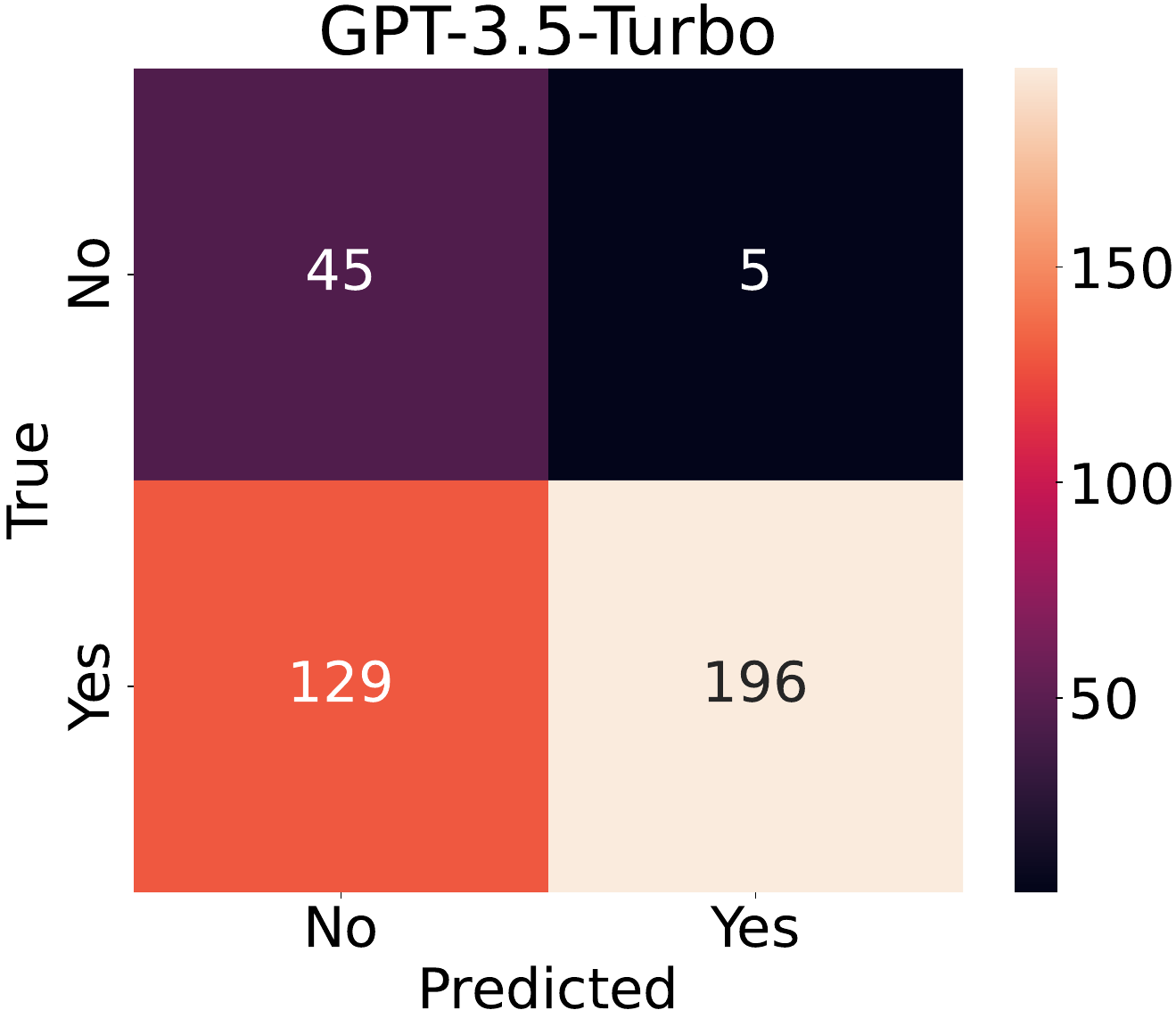}
\includegraphics[width=0.325\textwidth]{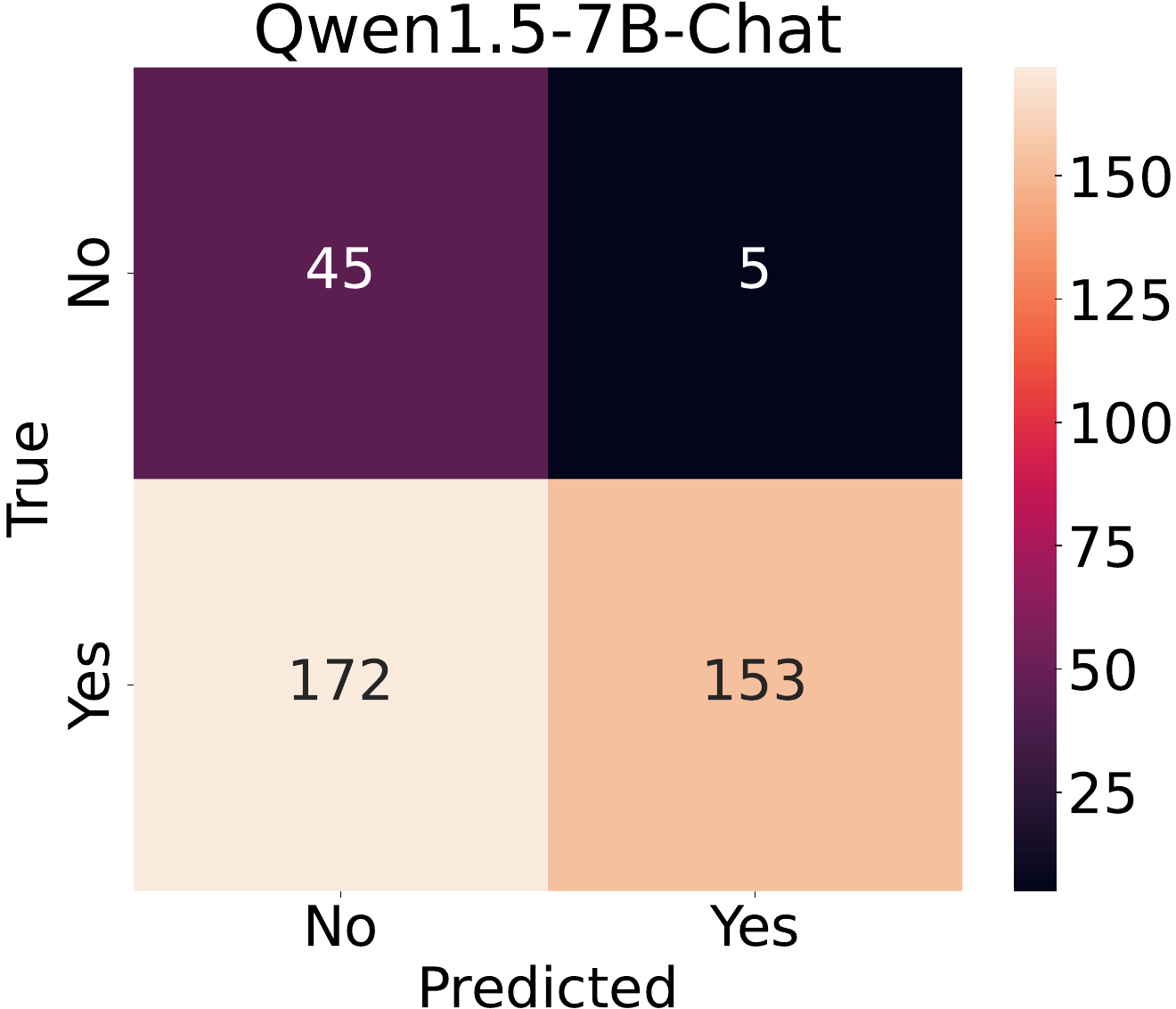}

\caption{Confusion matrices for the best models of each family (\textsc{ECoh-1.8B-en}, GPT-3.5-Turbo and \textsc{Qwen1.5-7B-Chat}) on FED-turn.}
\label{fig:cm}
\end{figure*}

From Table \ref{tab:fed_res}, we draw several conclusions. Firstly, when looking at the correlation metric, we see that the performance gap between random response-trained models and \textsc{ECoh} is much smaller. This is mainly due to the older chatbots models used for FED-turn -- \textbf{Meena} \citep{DBLP:journals/corr/abs-2001-09977} and \textbf{Mitsuku}\footnote{\href{https://medium.com/pandorabots-blog/mitsuku-wins-loebner-prize-2018-3e8d98c5f2a7}{Mitsuku blogpost}} -- being more likely to output irrelevant and non-specific responses that mimic random response selection.

Secondly, we note that our finetuning is still useful for detecting coherence issues on FED, since, overall, \textsc{ECoh} outperforms the corresponding parent model (e.g., \textsc{ECoh-4B} vs \textsc{Qwen1.5-4B}) on Relevance. However, our multilingual models underperform against the English-specific finetuning (with the exception of the 4B model) which could be explained by FED being exclusively in English. 

Finally, despite GPT-4 not being a perfect evaluator (low correlation in FED), we assume the scores we obtain for FED remain comparable to previous experiments. With that in mind, we find that the explanation quality is overall lower for \textsc{ECoh} and the teacher model, GPT-3.5-Turbo when compared to \textsc{GenResCoh} (for instance, we report an absolute drop of 1.01 for GPT-3.5-Turbo). In contrast, the \textsc{Qwen1.5-Chat} models' explanation quality remains almost unchanged when compared to \textsc{GenResCoh}. As a result, \textsc{ECoh} models achieve less impressive results when compared to their parent models. When comparing models of the same size up until 4B, we do see some improvement in explanation quality. However, our 4B model fails to outperform the 7B model, both in terms of correlation and explanation quality. For the correlation, we believe this is due to the parent model (\textsc{Qwen1.5-Chat-4B}) having low predictive performance -- if we look at the 1.8B models, they yield better correlations. For the explanation quality, we note that the teacher model, GPT-3.5, has also lower results than expected.

However, it is important to acknowledge that these models are tailored towards the binary classification of coherence. As such, we also present results for FED-turn with a binary mapping. In this case, we consider a response to be relevant when the majority of the annotators rate the response as fluent. We present the confusion matrices for the best models of each family in Figure \ref{fig:cm}. Here, we note that \textsc{ECoh-1.8B-en} is more likely to say a response is coherent, incurring slightly more False Positives than the other models (17 against 5), but also lower False Negatives (83 against >129).

\subsection{Error analysis}

In order to determine limitations and weaknesses with using \textsc{ECoh}, we conducted a thorough analysis of all errors in \textsc{GenResCoh-PC} and FED. We summarise significant findings below.

\begin{table}[ht]
\centering
\small
\begin{tabularx}{0.48\textwidth}{ X }
\toprule
\textbf{Context:}\\
\textbf{A:} Hi! \\
\textbf{B:} hi\\
\textbf{A:} what are you up to?\\
\textbf{B:} code refactoring. you?\\
\textbf{A:} me? just chilling out at work. what is code refactoring?\\\midrule
\textbf{Response. B:} good question. I don't even know what I am doing\\
Annotations: \hspace{0.2cm} \textbf{Relevant:} 1.6 \hspace{1cm} \textbf{Overall:} 3.0 \\\midrule
\cellcolor[HTML]{FDEDEC}\textbf{Evaluation:} \textit{"The response contradicts B's previous statement about code refactoring. The answer is No."}\\
\bottomrule
\end{tabularx}
\caption{Example prediction by \textsc{ECoh-1.8B-ml} on FED-Turn. Here, both speakers are human.}
\label{tab:cherry_negative1}
\end{table}

\paragraph{Humanlike Responses} One of the issues we found when analysing incorrect predictions is that the model sometimes rates human responses as incoherent when they are not "helpful" or do not directly respond to the context, which is not the expected behaviour with chatbot responses. For instance, Table \ref{tab:cherry_negative1} presents a cherry-picked example where \textsc{ECoh} incorrectly rates a human response from FED-Turn as incoherent. We believe such issues could be mitigated by including some portion of coherent human responses from DailyDialog++ during training. However, we also found similar issues with GPT-3.5-Turbo and \textsc{Qwen}, indicating this type of error is transversal to many current LLM-based evaluators.

\begin{table}[ht]
\centering
\small
\begin{tabularx}{0.48\textwidth}{ X }
\toprule
\textbf{Context:}\\
\textbf{A:} Che grande giornata! \textit{What a great day!}\\
\textbf{B:} Si. Che ne dici di uscire e Dioerti il sole sull'erba? \textit{Yes. How about we go out and enjoy the sun on the grass?}\\
\textbf{A:} Grande, andiamo! \textit{Great, let's go!}\\
\textbf{B:} Ehi, tesoro, credo di aver avuto un colpo di calore per essere stato tutto il giorno al sole. Sono così rilassato. Mi sembra di essere in un altro mondo. \textit{Hey, honey, I think I had heatstroke from being in the sun all day. I'm so relaxed. I feel like I'm in another world.}\\\midrule
\textbf{Response. A:} Oh no, forse dovremmo trovare un posto all'ombra o rientrare in casa. Come ti senti adesso? \textit{Oh no, maybe we should find a shady spot or go indoors. How do you feel now?}\\\midrule
\cellcolor[HTML]{ECF9E9}\textbf{Reference:} \textit{"The response is considerate and logical, offering a solution to avoid further sun exposure due to the heatstroke B mentioned, and it also shows concern for B’s current well-being. The answer is Yes."}\\\midrule
\cellcolor[HTML]{FDEDEC}\textbf{Evaluation:} \textit{"The response contradicts B's statement about feeling relaxed and suggests going back indoors. The answer is No."}\\
\bottomrule
\end{tabularx}
\caption{Example prediction by \textsc{ECoh-1.8B-ml} on \textsc{GenResCoh-DD-test}. Translation provided in \textit{italic}.}
\label{tab:cherry_negative2}
\end{table}

\paragraph{Commonsense Reasoning} Another issue we found recurrent, especially with the smaller models, is the limited nature of their commonsense reasoning. We hypothesise that this is a capability that smaller models struggle with, and this is reflected in their evaluation capabilities. We see an example of this is Table \ref{tab:cherry_negative2}, where the model fails to understand that sun exposure is mitigated by going back indoors.

\section{Related Work}

\subsection{Metrics for Dialogue Evaluation}

Statistic-based metrics such as BLEU \citep{papineni-etal-2002-bleu}, ROUGE \cite{lin2004rouge}, and METEOR \citep{banerjee-lavie-2005-meteor}, are a popular choice for dialogue evaluation because they are simple to calculate and lightweight. However, since they assume valid responses have significant word-overlap with the ground truth, their correlations with human judgements annotations are very low \citep{liu-etal-2016-evaluate} due to the one-to-many nature of dialogues. Additionally, they cannot be used to evaluate models whenever a gold-response is not available. 

Consequently, learned metrics were proposed. The typical approach was to finetune pretrained encoder models using positive and negative samples targeting different quality aspects such as fluency and relevance \citep{mehri-eskenazi-2020-usr,phy-etal-2020-deconstruct,sai-etal-2020-improving,mendonca-etal-2022-qualityadapt,zhao}. Other approaches used graph representations to model dialogue interactions explicitly \cite{huang-etal-2020-grade, zhang-etal-2021-dynaeval}. 

With the introduction of LLMs in a wide range of NLP tasks, most recent approaches leverage them for dialogue evaluation. \textsc{G-Eval} \citep{liu-etal-2023-g} uses GPT-3.5-Turbo and GPT-4 for the evaluation of generation models using a "Chain-Of-Thoughts" step and a scoring function based on return token probabilities. \textsc{LLM-Eval} \citep{lin-chen-2023-llm} is a single-prompt-based evaluation method that leverages a unified evaluation schema to cover multiple dimensions of conversation quality in a forward pass. \textsc{DialEvalML} \citep{mendonca-etal-2023-simple} combines encoder-based models and direct prompting and score extraction from GPT-3.5-Turbo. \textsc{XDial-Eval} \citep{zhang-etal-2023-xdial} probes the evaluation capabilities of several open source LLMs agaisnt GPT-3.5-Turbo \citep{ouyang2022training}, and also finetunes them with dialogue data. To the best of our knowledge, this is the first work that conducts supervised learning of LLMs for the task of dialogue evaluation.

\subsection{Dataset Generation}

There are several studies that propose augmentation and synthetic generation approaches to scale dataset sizes that target commonsense reasoning \citep{bhagavatula-etal-2023-i2d2, wang-etal-2023-scott}, summarisation \citep{jung2024impossible}, and dialogues \citep{chen-etal-2023-places,kim-etal-2023-soda} for training purposes.

For dialogue evaluation in particular, most metrics are finetuned using self-supervised data \citep{mehri-eskenazi-2020-usr,phy-etal-2020-deconstruct,yeh-etal-2021-comprehensive,mendonca-etal-2023-towards}. The most widely used approach is to select positive samples consisting of the ground truth response, and negative responses from randomly drawn dialogues. \citet{ghazarian-etal-2022-deam} relies on Abstract Meaning Representation (AMR) to apply semantic-level manipulations to existing responses. Our work, in comparison, leverages a strong LLM to generate new incoherent responses at scale.

\section{Conclusions}

This paper presents \textsc{GenResCoh}, a large scale collection of positive and negative responses and corresponding explanations covering several languages. \textsc{GenResCoh} is generated from XDailyDialog and XPersona using state-of-the-art LLMs, which better matches the responses seen by contemporary chatbots. With this dataset, we train a family of evaluators we call \textsc{ECoh}. Our smallest model (0.5B) is able to achieve similar performance to that of the teacher model (GPT-3.5-Turbo), despite being much smaller.

Despite this good performance, we note some limitations when using \textsc{ECoh}, especially when evaluating human responses and/or responses that require more robust commonsense reasoning. Whilst we argue that including more data that targets commonsense and human responses, or even training a larger model could mitigate these issues, since we include an explanation in the predictions, one could still use our evaluators for an initial evaluation screening and escalate to a human evaluator if necessary.

\section{Limitations}

\paragraph{Reduced Language Selection} Our work is only evaluated in English, German, Italian, French and Chinese. This limitation stems in part from the upstream dialogue dataset (XDailyDialog) only covering 4 high resource languages. Whilst XPersona does contain additional languages, we were limited to only including French as unseen language due to annotator and resource limitations. 

\paragraph{Generation} Generating synthetic data from LLMs might surface or even amplify harmful content within these models. In particular, the choice of a single LLM to generate the responses may induce distribution biases. We identify in Section \ref{sec:negeval_dataset} the reduced lexical diversity of generated responses from GPT-3.5-Turbo when compared to humans. Furthermore, our limited analysis shows that our model sometimes struggles with rating human responses. As such, the generated negative samples may also not accurately represent all coherence issues LLM-based generators typically exhibit. Future investigation may look into producing a systematic quality analysis of a more diverse pool of LLMs, which could inform more faithful generation of negative responses.

\paragraph{FED as a turn level coherence benchmark} For most dialogue evaluation benchmarks, coherence annotations are conducted at the dialogue level and do not pinpoint the exact response that triggers incoherence \cite{yeh-etal-2021-comprehensive}. As such, we opted with benchmarking \textsc{ECoh} on FED-turn relevance annotations, which is a typically used benchmark for dialogue evaluation. Despite relevance and coherence being different quality aspects, we note that a) all irrelevant responses lack, by definition, coherence; b) we found that the vast majority of relevant responses on FED are also coherent. Nevertheless, we acknowledge the limitations of using FED-turn as a turn level coherence evaluation benchmark, namely due to its lack of relevant but incoherent responses. 

\section{Ethical Considerations}

\paragraph{Culture-specific conversational norms} We acknowledge that the definition of dialogue quality is a diverse, culturally informed concept. We attempt to reduce the English-centric bias in the generation by leaving the LLM to generate without English reference constraints. However it is possible the generation still conforms to English definitions of coherence given its pretraining and instruction tuning data is more than likely over represented by English text. Furthermore, the examples provided in the prompt, and the dialogues themselves, despite being validated by expert linguists, are still based on English dialogues. As such, users of our model should take extra care when evaluating responses in languages that are known to deviate substantially from English-centric notions of coherence.

\paragraph{Annotations} The post-editing of the prompts and the manual validation of GPT-4 generations was partially conducted by volunteer annotators, and paid workers that have a fair wage according to their location.

\section*{Acknowledgments}

This research was supported by the Portuguese Recovery and Resilience Plan through project C645008882-00000055 (Responsible.AI) and by national funds through \textit{Fundação para a Ciência e a Tecnologia} (FCT) with references PRT/BD/152198/2021 and DOI: 10.54499/UIDB/50021/2020.

\bibliography{anthology,custom}

\appendix

\section{Dataset Curation}

\subsection{Generation}
\label{sec:app_generate}

\paragraph{Prompt} The prompt, which is shared for the development and test set is presented in Table \ref{tab:prompt_negeval}. For each language, we translate the example dialogues and responses using Google Translate\footnote{\url{https://translate.google.com}} and manually validate the full prompt with the expert linguists, ensuring the explanation is accurate for the translated response.

\begin{table}[ht]
\small
\centering
\begin{tabularx}{0.48\textwidth}{ X }
\cellcolor[HTML]{eeeeee}Given the dialog, generate a good and a bad response. In particular, the bad response should have issues that reduce its quality in terms of coherence, such as contradictions, logical inconsistencies, etc. Output the responses, together with a small explanation of the response using the following json format:\\\vspace{0.2cm}
\cellcolor[HTML]{eeeeee} \{"good\_response": "..." , "good\_explanation": "...", "bad\_response": "...", "bad\_explanation": "..."\} \\\vspace{0.001cm}
\cellcolor[HTML]{eeeeee}Examples:\\\vspace{0.001cm}
\cellcolor[HTML]{eeeeee}Dialogue: A: Have you figured out where you want to transfer to? B: I can't think of where to go. A: Where would you like to go to school?\\\vspace{0.1cm}
\cellcolor[HTML]{eeeeee}Output: \{"good\_response": "B: Well, It is not yet decided, but maybe in the east coast." , "good\_explanation": "The response acknowledges the question and provides a region.", "bad\_response": "B: Do you think that I can get married after school?", "bad\_explanation" : "The response does not acknowledge the prior question."\} \\\vspace{0.2cm}
\cellcolor[HTML]{eeeeee}Dialogue: A: You look so tan and healthy! B: Thanks. I just got back from summer camp A: How was it ? B: Great. I got to try so many things for the first time.\\\vspace{0.1cm}
\cellcolor[HTML]{eeeeee}Output: \{"good\_response": "A: I wish I could go to summer camp too. I'm so bored at home.", "good\_explanation": "The response acknowledges the positive emotions displayed and contrasts it with their own perspective of summer break.", "bad\_response": "A: Did you eat while you where there? You look frail.", "bad\_explanation": "The response contradicts the earlier statement indicating they were healthy."\} \\\vspace{0.2cm}
\cellcolor[HTML]{eeeeee}Dialogue:\\
\cellcolor[HTML]{eeeeee}\hspace{0.18\textwidth}\textbf{[Dialogue]}
\end{tabularx}
\caption{Response generation instruction template.}
\label{tab:prompt_negeval}
\end{table}

\paragraph{Cost} In total, our generation using a mixture of GPT-3.5-Turbo and GPT-4 amounted to approximately 300\$ USD, with 2/3 of the budget allocated to the generation of test set responses for both XDailyDialog and XPersona.

\subsection{Additional Examples}
\label{sec:app_ex}

Table \ref{tab:negeval_ex_zh}, \ref{tab:negeval_ex_de}, \ref{tab:negeval_ex_it} showcase additional examples resulting from GPT-4 response generation for XDailyDialog. Additionally, an example from XPersona for French is in Table \ref{tab:negeval_ex_fr}. For convenience, we provide MT in \textit{italic}.\footnote{These were not checked nor post-edited.}

\begin{CJK*}{UTF8}{gbsn}
\begin{table}[ht]
\centering
\small
\begin{tabularx}{0.48\textwidth}{ X }
\toprule
\textbf{Context:}\\
\textbf{A:} 你好，苗莉，你要去哪里？\textit{Hello, Miao Li, where are you going?}\\
\textbf{B:} 你好，我要去商店买些水果。\textit{Hello, I'm going to the store to buy some fruit.}\\\midrule
\textbf{Positive Response. A:} 哦，那真好，要买哪些水果 \textit{Oh, that's great. What kind of fruit to buy?} \\
\cellcolor[HTML]{ECF9E9}\textit{"The response naturally continues the conversation by showing interest in the specific types of fruits B is planning to buy."}\\\midrule
\textbf{Negative Response. A:} 为什么你现在才告诉我你要去商店？ \textit{Why are you just telling me now that you're going to the store?}\\
\cellcolor[HTML]{FDEDEC}\textit{"The response introduces a sense of confrontation and confusion that wasn't present in the prior conversation. It suggests an expectation that B should have informed A earlier, which is not supported by the dialogue."}\\
\bottomrule
\end{tabularx}
\caption{Example test set annotation for XDailyDialog for Chinese-\textsc{zh}. MT in \textit{Italic}.}
\label{tab:negeval_ex_zh}
\end{table}
\end{CJK*}

\begin{table}[h!]
\centering
\small
\begin{tabularx}{0.48\textwidth}{ X }
\toprule
\textbf{Context:}\\
\textbf{A:} Wo wollen Sie hin? \textit{Where do you want to go?} \\
\textbf{B:} Ich fahre ins Krankenhaus. \textit{I'm going to the hospital.}\\
\textbf{A:} Steigen Sie ein, ich bringe Sie hin. \textit{Get in, I'll take you there.}\\
\textbf{B:} Wie viel wird es kosten? \textit{How much will it cost?}\\
\textbf{A:} Es kostet Sie 10 Dollar. \textit{It will cost you 10 dollars.}\\
\textbf{B:} Übrigens, wie spät ist es? \textit{By the way, what time is it?}\\\midrule
\textbf{Positive Response. A:} Es ist jetzt halb drei nachmittags. \textit{It's now half past two in the afternoon.} \\
\cellcolor[HTML]{ECF9E9}\textit{"The response directly and appropriately answers B's question about the current time, maintaining the flow of the conversation."}\\\midrule
\textbf{Negative Response. A:} Oh, ich habe keine Uhr, also weiß ich nie, wie spät es ist. \textit{Oh, I don't have a watch, so I never know what time it is.}\\
\cellcolor[HTML]{FDEDEC}\textit{"This response is inconsistent because as a driver, it would be expected that A has a way to tell the time, especially when their job involves scheduling. It contradicts the setting where a driver should be aware of the time to manage their routes and appointments."}\\
\bottomrule
\end{tabularx}
\caption{Example test set annotation for XDailyDialog for German-\textsc{de}. MT in \textit{Italic}.}
\label{tab:negeval_ex_de}
\end{table}

\begin{table}[h!]
\centering
\small
\begin{tabularx}{0.48\textwidth}{ X }
\toprule
\textbf{Context:}\\
\textbf{A:} Sto cercando il vecchio carillon. \textit{I'm looking for the old music box.} \\
\textbf{B:} Sei arrivato nel posto giusto, ci sono dei requisiti cronologici? \textit{You've come to the right place, are there any chronological requirements?}\\
\textbf{A:} Sarebbe fantastico se ci fosse un 1920 qui. \textit{It would be great if there was a 1920's here.}\\
\textbf{B:} Ne abbiamo ricevuto solo 1 ieri e ora ce ne sono 6 qui. \textit{We only received 1 yesterday and now there are 6 here.}\\\midrule
\textbf{Positive Response. A:} Davvero? Potrei vederli per scegliere il migliore? \textit{Really? Could I see them to choose the best one?}\\
\cellcolor[HTML]{ECF9E9}\textit{"The response shows interest and eagerness to see the selection which logically follows from the information provided about the availability of the items."}\\\midrule
\textbf{Negative Response. A:} Oh, non mi interessano i carillon degli anni '20. \textit{Oh, I don't care about 1920s music boxes.}\\
\cellcolor[HTML]{FDEDEC}\textit{"The response is inconsistent as it contradicts the earlier expressed interest in a 1920s carillon, making it illogical and irrelevant to the previous exchange."}\\
\bottomrule
\end{tabularx}
\caption{Example test set annotation for XDailyDialog for Italian-\textsc{it}. MT in \textit{Italic}.}
\label{tab:negeval_ex_it}
\end{table}

\begin{table}[h!]
\centering
\small
\begin{tabularx}{0.48\textwidth}{ X }
\toprule
\textbf{Context:}\\
\textbf{A:} Salut! êtes-vous déjà monté à cheval? \textit{Hi! have you ever ridden a horse?} \\
\textbf{B:} Oui! Ça fait partie de mon traitement contre le cancer. J'ai commencé un nouveau travail. Toi? \textit{Yes ! It’s part of my cancer treatment. I started a new job. You?}\\
\textbf{A:} J'ai entendu dire que couper la viande de votre alimentation aide à combattre le cancer. \textit{I heard cutting meat from your diet helps fight cancer.}\\\midrule
\textbf{Positive Response. B:} C'est intéressant, je devrais peut-être envisager cela aussi. Comment ça se passe pour toi? \textit{That's interesting, maybe I should consider that too. How is it going for you?} \\
\cellcolor[HTML]{ECF9E9}\textit{"The response shows interest in the advice given by A and seeks more personal insight, which maintains the flow of the conversation and shows engagement."}\\\midrule
\textbf{Negative Response. B:} Oui, et c'est pourquoi je mange plus de viande maintenant pour rester en forme. \textit{Yes, and that's why I eat more meat now to stay in shape.}\\
\cellcolor[HTML]{FDEDEC} \textit{"The response is contradictory because A just mentioned that cutting meat helps fight cancer, yet B responds by saying they eat more meat to stay fit, which directly contradicts the advice given."}\\
\bottomrule
\end{tabularx}
\caption{Example test set annotation for Xpersona in French-\textsc{fr}. MT in \textit{Italic}.}
\label{tab:negeval_ex_fr}
\end{table}

\subsection{Manual validation}
\label{sec:app_validate}
\begin{figure*}[h]
    \centering
    \frame{\includegraphics[width=\linewidth]{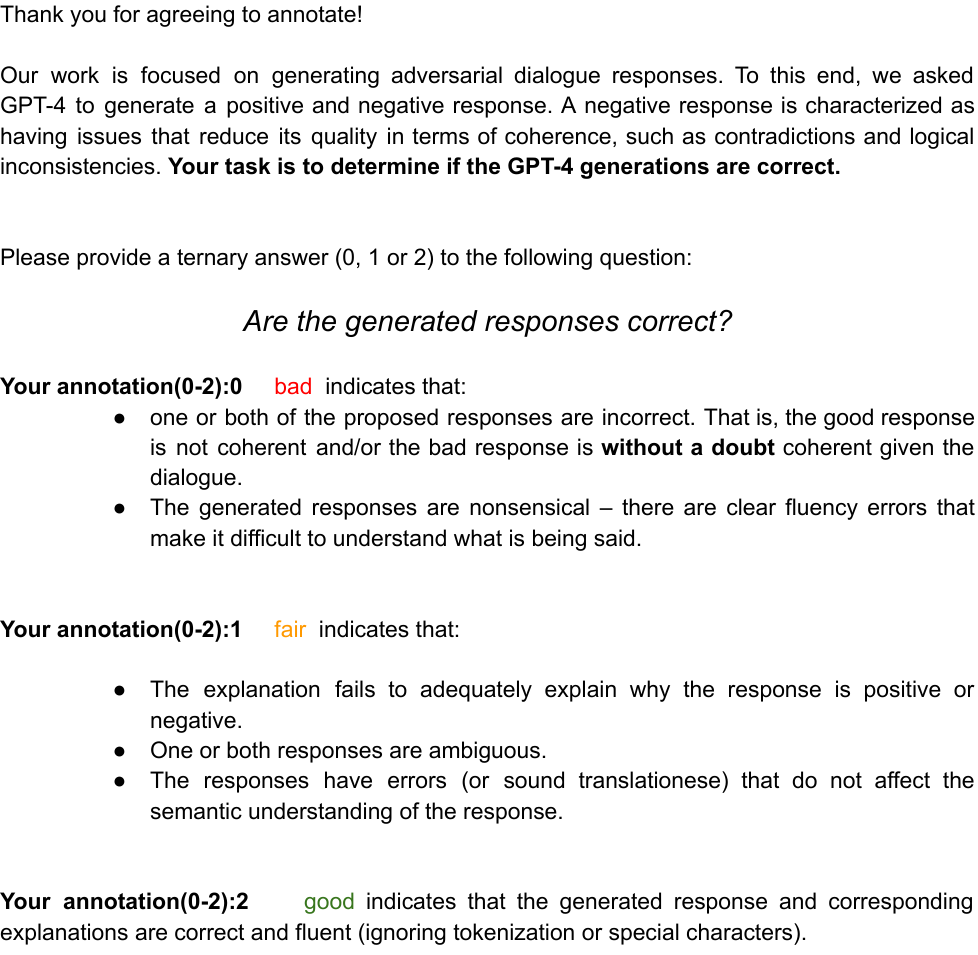}}
    \caption{GPT-4 response validation guidelines.}
    \label{fig:negeval_guidelines}
\end{figure*}

For the manual validation step, a single annotator for each language is recruited to validate the response and corresponding explanation. A total of 100 randomly selected examples from the test set (generated by GPT-4) were sampled, per language, for this validation. We consider a response to be appropriate if the annotation is 1 or above (fair). The full guidelines provided to the expert annotators are presented in Figure \ref{fig:negeval_guidelines}.

\section{Implementation Details}
\label{sec:exp_details}

\subsection{\textsc{NSP-RoBERTa}}

We use the \textsc{RoBERTa} large encoder model downloaded from HuggingFace \footnote{\url{huggingface.co/roberta-large}} for all experiments. We train a regression model on a single RTX A6000 GPU using the following sampling strategy: Given a fixed context from DailyDialog, \textbf{positive} responses are drawn directly from the same dialog; \textbf{negative} responses are randomly selected and a token coverage test discards semantically similar sentences. In total, 89,707/38,449 datapoints were obtained after processing. 

A token representing the speaker was added for each turn, and a history length of 3 turns was used. We applied a regression head consisting of a 2-layer MLP with a hidden size of 1024 and a hyperbolic tangent function as activation for prediction. All parameters were trained/finetuned using Adam optimizer \cite{DBLP:journals/corr/KingmaB14}, using a learning rate of 3e-6 and were trained for 3 epochs using a batch size of 16. Evaluation was conducted every 1,000 steps. The best performing model on the evaluation set was selected for testing. 

\subsection{\textsc{ECoh}}

We train the \textsc{ECoh} models on a mixture of A100 80GB and RTX A6000 GPUs (depending on model size). We finetune using Huggingface Transfomers and PEFT\footnote{\url{huggingface.co/docs/peft}} for a 3 epochs for the English model and 1 epoch for the multilingual model with early stopping. We finetune from the base \textsc{Qwen1.5-Chat} models (full precision) using LoRA \citep{hu2021lora}, with $r=8$, $\alpha=32$ and dropout set to 0.1. Gradient accumulation steps is set to 4 with a learning rate of $1e-4$. Batch size was set to maximize VRAM consumption, ranging from 2 up to 8 per device. 

For inference, we follow \textsc{Qwen1.5-Chat} inference code\footnote{\url{github.com/QwenLM/Qwen1.5}}, which generates responses using sampling with a temperature of 1, repetition penalty of 1.1, and top $p$ set to 0.8.

\section{Additional Results}
\label{sec:app_res}

This appendix presents the individual results for zero shot, 1 shot with english example, 1 shot with target language example and the finetuned \textsc{ECoh} models for each individual language for for \textsc{GenResCoh-DD-test}, sorted by model size -- 0.5B (Table \ref{tab:dd_test_0.5b}), 1.8B and 4B (Table \ref{tab:dd_test_4b}) and 7B and GPT-3.5-Turbo (Table \ref{tab:dd_test_7b}). Table \ref{tab:pc_test} presents the results for \textsc{GenResCoh-PC-test}.

\begin{table}[h]
\small
\centering
\begin{tabular}{lcccc}
\toprule
\textbf{Model}              & $\rho_{pb}$ & \textbf{F1}    & \textbf{BLEU}  & \textbf{GPT-4} \\\midrule

\multicolumn{5}{l}{\textsc{Qwen1.5-0.5B-Chat-0shot}}\\\cmidrule{1-1}
\hspace{1.5cm}\textsc{en}   & .2141   & .45 & 2.15  & -     \\
\hspace{1.5cm}\textsc{de}   & .1382   & .39 & 1.80  & -     \\
\hspace{1.5cm}\textsc{it}   & .1695   & .41 & 1.99  & -     \\
\hspace{1.5cm}\textsc{zh}   & .1977   & .44 & 1.88  & -     \\\midrule

\multicolumn{5}{l}{\textsc{Qwen1.5-0.5B-Chat-1shot-en}}\\\cmidrule{1-1}
\hspace{1.5cm}\textsc{en}   & .2662   & .60 & 2.69  & 2.12\tiny{\textpm1.20}     \\
\hspace{1.5cm}\textsc{de}   & .1967   & .55 & 2.19  & -     \\
\hspace{1.5cm}\textsc{it}   & .2210   & .55 & 2.47  & -     \\
\hspace{1.5cm}\textsc{zh}   & .2361   & .59 & 2.25  & -     \\\midrule

\multicolumn{5}{l}{\textsc{Qwen1.5-0.5B-Chat-1shot-lang}}\\\cmidrule{1-1}
\hspace{1.5cm}\textsc{en}   & .2662   & .60 & 2.69  & 2.12\tiny{\textpm1.20}     \\
\hspace{1.5cm}\textsc{de}   & .1870   & .59 & 4.35  & 1.40\tiny{\textpm0.77}     \\
\hspace{1.5cm}\textsc{it}   & .1567   & .56 & 4.46  & 1.84\tiny{\textpm1.24}     \\
\hspace{1.5cm}\textsc{zh}   & .2803   & .64 & 3.70  & 2.08\tiny{\textpm1.25}     \\\midrule

\multicolumn{5}{l}{\textsc{ECoh-0.5B-en}}\\\cmidrule{1-1}
\hspace{1.5cm}\textsc{en}   & .8995   & .95 & 19.34  & 4.24\tiny{\textpm1.01}    \\
\hspace{1.5cm}\textsc{de}   & .6407   & .79 & 14.42  & 3.28\tiny{\textpm1.72}     \\
\hspace{1.5cm}\textsc{it}   & .7035   & .84 & 14.41  & 3.68\tiny{\textpm1.57}     \\
\hspace{1.5cm}\textsc{zh}   & .8587   & .93 & 15.89  & 3.92\tiny{\textpm1.41}     \\\midrule

\multicolumn{5}{l}{\textsc{ECoh-0.5B-ml}}\\\cmidrule{1-1}
\hspace{1.5cm}\textsc{en}   & .9174   & .96 & 19.34  & 4.08\tiny{\textpm1.15}    \\
\hspace{1.5cm}\textsc{de}   & .8749   & .94 & 14.42  & 4.04\tiny{\textpm1.30}     \\
\hspace{1.5cm}\textsc{it}   & .8565   & .93 & 14.41  & 3.48\tiny{\textpm1.66}     \\
\hspace{1.5cm}\textsc{zh}   & .9038   & .95 & 15.89  & 4.20\tiny{\textpm1.32}     \\\midrule

\end{tabular}
\caption{Reported results for \textsc{GenResCoh-DD-test} for the 0.5B models. $\rho_{pb}$ denotes Point Biserial Correlation.  All correlation results are $p<0.05$.}
\label{tab:dd_test_0.5b}
\end{table}
\begin{table}[h]
\small
\centering
\begin{tabular}{lcccc}
\toprule
\textbf{Model}              & $\rho_{pb}$ & \textbf{F1}    & \textbf{BLEU}  & \textbf{GPT-4} \\\midrule

\multicolumn{5}{l}{\textsc{Qwen1.5-1.8B-Chat-0shot}}\\\cmidrule{1-1}
\hspace{1.5cm}\textsc{en}   & .4765   & .67 & 2.14  & -     \\
\hspace{1.5cm}\textsc{de}   & .2663   & .49 & 1.80  & -     \\
\hspace{1.5cm}\textsc{it}   & .3207   & .54 & 1.85  & -     \\
\hspace{1.5cm}\textsc{zh}   & .4047   & .62 & 1.95  & -     \\\midrule

\multicolumn{5}{l}{\textsc{Qwen1.5-1.8B-Chat-1shot-en}}\\\cmidrule{1-1}
\hspace{1.5cm}\textsc{en}   & .5473   & .74 & 2.70  & 2.36\tiny{\textpm1.38}     \\
\hspace{1.5cm}\textsc{de}   & .4413   & .68 & 2.50  & -     \\
\hspace{1.5cm}\textsc{it}   & .4430   & .68 & 2.76  & -     \\
\hspace{1.5cm}\textsc{zh}   & .5652   & .76 & 2.46  & -     \\\midrule

\multicolumn{5}{l}{\textsc{Qwen1.5-1.8B-Chat-1shot-lang}}\\\cmidrule{1-1}
\hspace{1.5cm}\textsc{en}   & .5473   & .74 & 2.70  & 2.36\tiny{\textpm1.38}     \\
\hspace{1.5cm}\textsc{de}   & .4680   & .71 & 2.38  & 2.08\tiny{\textpm1.22}     \\
\hspace{1.5cm}\textsc{it}   & .4536   & .72 & 3.03  & 2.28\tiny{\textpm1.31}     \\
\hspace{1.5cm}\textsc{zh}   & .6160   & .79 & 2.21  & 2.12\tiny{\textpm1.27}     \\\midrule

\multicolumn{5}{l}{\textsc{ECoh-1.8B-en}}\\\cmidrule{1-1}
\hspace{1.5cm}\textsc{en}   & .9227   & .96 & 20.15  & 4.62\tiny{\textpm0.85}    \\
\hspace{1.5cm}\textsc{de}   & .7432   & .86 & 16.11  & 4.00\tiny{\textpm1.40}     \\
\hspace{1.5cm}\textsc{it}   & .7381   & .86 & 15.57  & 3.76\tiny{\textpm1.59}     \\
\hspace{1.5cm}\textsc{zh}   & .8926   & .95 & 17.35  & 4.12\tiny{\textpm1.33}     \\\midrule

\multicolumn{5}{l}{\textsc{ECoh-1.8B-ml}}\\\cmidrule{1-1}
\hspace{1.5cm}\textsc{en}   & .9327   & .97 & 20.08  & 4.32\tiny{\textpm1.31}    \\
\hspace{1.5cm}\textsc{de}   & .8859   & .94 & 15.92  & 4.04\tiny{\textpm1.50}     \\
\hspace{1.5cm}\textsc{it}   & .8732   & .94 & 15.23  & 4.04\tiny{\textpm1.40}     \\
\hspace{1.5cm}\textsc{zh}   & .9159   & .96 & 17.88  & 4.56\tiny{\textpm0.92}     \\\midrule\midrule

\multicolumn{5}{l}{\textsc{Qwen1.5-4B-Chat-0shot}}\\\cmidrule{1-1}
\hspace{1.5cm}\textsc{en}   & .7365   & .86 & 3.57  & -     \\
\hspace{1.5cm}\textsc{de}   & .6501   & .82 & 3.49  & -     \\
\hspace{1.5cm}\textsc{it}   & .6275   & .81 & 3.55  & -     \\
\hspace{1.5cm}\textsc{zh}   & .7138   & .85 & 3.48  & -     \\\midrule

\multicolumn{5}{l}{\textsc{Qwen1.5-4B-Chat-1shot-en}}\\\cmidrule{1-1}
\hspace{1.5cm}\textsc{en}   & .6163   & .79 & 4.08  & 3.56\tiny{\textpm1.50}     \\
\hspace{1.5cm}\textsc{de}   & .5764   & .78 & 4.06  & -     \\
\hspace{1.5cm}\textsc{it}   & .5728   & .78 & 4.34  & -     \\
\hspace{1.5cm}\textsc{zh}   & .5400   & .73 & 3.80  & -     \\\midrule

\multicolumn{5}{l}{\textsc{Qwen1.5-4B-Chat-1shot-lang}}\\\cmidrule{1-1}
\hspace{1.5cm}\textsc{en}   & .6163   & .79 & 4.08  & 3.56\tiny{\textpm1.50}     \\
\hspace{1.5cm}\textsc{de}   & .5754   & .79 & 7.72  & 2.64\tiny{\textpm1.73}     \\
\hspace{1.5cm}\textsc{it}   & .5269   & .75 & 13.78 & 3.52\tiny{\textpm1.69}     \\
\hspace{1.5cm}\textsc{zh}   & .6213   & .80 & 7.04  & 3.60\tiny{\textpm1.50}     \\\midrule

\multicolumn{5}{l}{\textsc{ECoh-4B-en}}\\\cmidrule{1-1}
\hspace{1.5cm}\textsc{en}   & .9464   & .97 & 20.66  & 4.60\tiny{\textpm0.91}    \\
\hspace{1.5cm}\textsc{de}   & .8980   & .95 & 17.17  & 4.64\tiny{\textpm0.77}     \\
\hspace{1.5cm}\textsc{it}   & .8982   & .95 & 16.42  & 3.92\tiny{\textpm1.19}     \\
\hspace{1.5cm}\textsc{zh}   & .9315   & .97 & 17.44  & 4.62\tiny{\textpm0.86}     \\\midrule

\multicolumn{5}{l}{\textsc{ECoh-4B-ml}}\\\cmidrule{1-1}
\hspace{1.5cm}\textsc{en}   & .9631   & .98 & 20.74  & 4.28\tiny{\textpm1.34}    \\
\hspace{1.5cm}\textsc{de}   & .9437   & .97 & 16.93  & 4.38\tiny{\textpm1.25}     \\
\hspace{1.5cm}\textsc{it}   & .9377   & .97 & 15.99  & 3.88\tiny{\textpm1.67}     \\
\hspace{1.5cm}\textsc{zh}   & .9520   & .98 & 18.52  & 4.34\tiny{\textpm1.00}     \\\midrule

\end{tabular}
\caption{Reported results for \textsc{GenResCoh-DD-test} for the 1.8B and 4B models. $\rho_{pb}$ denotes Point Biserial Correlation. All correlation results are $p<0.05$.}
\label{tab:dd_test_4b}
\end{table}

\begin{table}[h]
\small
\centering
\begin{tabular}{lcccc}
\toprule
\textbf{Model}              & $\rho_{pb}$ & \textbf{F1}    & \textbf{BLEU}  & \textbf{GPT-4} \\\midrule

\multicolumn{5}{l}{\textsc{Qwen1.5-7B-Chat-0shot}}\\\cmidrule{1-1}
\hspace{1.5cm}\textsc{en}   & .7490   & .86 & 4.30  & -     \\
\hspace{1.5cm}\textsc{de}   & .4868   & .66 & 4.74  & -     \\
\hspace{1.5cm}\textsc{it}   & .4302   & .61 & 4.90  & -     \\
\hspace{1.5cm}\textsc{zh}   & .6739   & .81 & 4.70  & -     \\\midrule

\multicolumn{5}{l}{\textsc{Qwen1.5-7B-Chat-1shot-en}}\\\cmidrule{1-1}
\hspace{1.5cm}\textsc{en}   & .8745   & .94 & 4.62  & 3.76\tiny{\textpm1.63}     \\
\hspace{1.5cm}\textsc{de}   & .7938   & .90 & 4.75  & -     \\
\hspace{1.5cm}\textsc{it}   & .7711   & .88 & 4.85  & -     \\
\hspace{1.5cm}\textsc{zh}   & .8210   & .91 & 4.37  & -     \\\midrule

\multicolumn{5}{l}{\textsc{Qwen1.5-7B-Chat-1shot-lang}}\\\cmidrule{1-1}
\hspace{1.5cm}\textsc{en}   & .8745   & .94 & 4.62  & 3.76\tiny{\textpm1.63}     \\
\hspace{1.5cm}\textsc{de}   & .7998   & .90 & 4.59  & 3.64\tiny{\textpm0.45}     \\
\hspace{1.5cm}\textsc{it}   & .6722   & .81 & 5.07  & 3.76\tiny{\textpm1.78}     \\
\hspace{1.5cm}\textsc{zh}   & .8208   & .91 & 5.07  & 4.28\tiny{\textpm1.14}     \\\midrule\midrule

\multicolumn{5}{l}{\textsc{GPT-3.5-Turbo-0shot}}\\\cmidrule{1-1}
\hspace{1.5cm}\textsc{en}   & .8592   & .93 & 4.92  & 4.58\tiny{\textpm1.04}    \\
\hspace{1.5cm}\textsc{de}   & .8218   & .91 & 5.47  & 4.66\tiny{\textpm0.85}     \\
\hspace{1.5cm}\textsc{it}   & .8102   & .90 & 5.44  & 4.36\tiny{\textpm1.41}     \\
\hspace{1.5cm}\textsc{zh}   & .8113   & .90 & 5.18  & 4.54\tiny{\textpm1.23}     \\\midrule

\end{tabular}
\caption{Reported results for \textsc{GenResCoh-DD-test} for the 7B models and GPT-3.5-Turbo. $\rho_{pb}$ denotes Point Biserial Correlation. All correlation results are $p<0.05$.}
\label{tab:dd_test_7b}
\end{table}

\begin{table}[h]
\small
\centering
\begin{tabular}{lcccc}
\toprule
\textbf{Model}              & $\rho_{pb}$ & \textbf{F1}    & \textbf{BLEU}  & \textbf{GPT-4} \\\midrule

\multicolumn{5}{l}{\textsc{Qwen1.5-7B-Chat-1shot}}\\\cmidrule{1-1}
\hspace{1.5cm}\textsc{en}   & .5787   & .75 & 3.28  & 4.28\tiny{\textpm1.45}     \\
\hspace{1.5cm}\textsc{fr}   & .4608   & .66 & 2.97  & 3.20\tiny{\textpm1.58}     \\
\hspace{1.5cm}\textsc{it}   & .6474   & .82 & 3.06  & 3.60\tiny{\textpm1.63}     \\
\hspace{1.5cm}\textsc{zh}   & .7630   & .88 & 4.40  & 3.92\tiny{\textpm1.32}     \\\midrule\midrule

\multicolumn{5}{l}{\textsc{ECoh-0.5B-ml}}\\\cmidrule{1-1}
\hspace{1.5cm}\textsc{en}   & .9021   & .95 & 17.71  & 3.90\tiny{\textpm1.32}     \\
\hspace{1.5cm}\textsc{fr}   & .8089   & .90 & 13.71  & 3.68\tiny{\textpm1.46}     \\
\hspace{1.5cm}\textsc{it}   & .8661   & .93 & 14.34  & 3.88\tiny{\textpm1.33}     \\
\hspace{1.5cm}\textsc{zh}   & .9260   & .96 & 15.84  & 3.80\tiny{\textpm1.38}     \\\midrule

\multicolumn{5}{l}{\textsc{ECoh-1.6B-ml}}\\\cmidrule{1-1}
\hspace{1.5cm}\textsc{en}   & .9043   & .95 & 18.17  & 4.44\tiny{\textpm1.12}     \\
\hspace{1.5cm}\textsc{fr}   & .8634   & .93 & 14.01  & 4.40\tiny{\textpm1.15}     \\
\hspace{1.5cm}\textsc{it}   & .8872   & .94 & 14.65  & 4.04\tiny{\textpm1.13}     \\
\hspace{1.5cm}\textsc{zh}   & .9390   & .97 & 16.35  & 4.16\tiny{\textpm1.28}     \\\midrule

\multicolumn{5}{l}{\textsc{ECoh-4B-ml}}\\\cmidrule{1-1}
\hspace{1.5cm}\textsc{en}   & .9443   & .97 & 18.81  & 4.40\tiny{\textpm1.04}    \\
\hspace{1.5cm}\textsc{fr}   & .9270   & .96 & 14.58  & 4.36\tiny{\textpm0.95}     \\
\hspace{1.5cm}\textsc{it}   & .9381   & .97 & 15.15  & 4.36\tiny{\textpm0.91}     \\
\hspace{1.5cm}\textsc{zh}   & .9700   & .98 & 16.79  & 4.40\tiny{\textpm0.96}     \\\midrule\midrule

\multicolumn{5}{l}{\textsc{GPT-3.5-Turbo-0shot}}\\\cmidrule{1-1}
\hspace{1.5cm}\textsc{en}   & .7767   & .89 & 4.71  & 4.08\tiny{\textpm1.38}    \\
\hspace{1.5cm}\textsc{fr}   & .7205   & .86 & 5.04  & 4.43\tiny{\textpm1.24}     \\
\hspace{1.5cm}\textsc{it}   & .8102   & .90 & 5.11  & 4.64\tiny{\textpm0.86}     \\
\hspace{1.5cm}\textsc{zh}   & .7452   & .87 & 3.89  & 4.76\tiny{\textpm0.72}     \\\midrule

\end{tabular}
\caption{Reported results for \textsc{GenResCoh-PC-test}. $\rho_{pb}$ denotes Point Biserial Correlation. All correlation results are $p<0.05$.}
\label{tab:pc_test}
\end{table}

\end{document}